\documentclass[11pt, letterpaper]{nyu}
\usepackage[authoryear,sort&compress,round]{natbib}
\usepackage[capitalize]{cleveref}
\usepackage{subcaption}

\newtoks\extrafooter
\newcommand{\setextrafooter}[1]{\extrafooter={#1}}

\title{YOR: Your Own Mobile Manipulator for Generalizable Robotics}

\correspondingauthor{}
\reportnumber{}
\websitedef{\href{https://yourownrobot.ai}{yourownrobot.ai}}

\author[1]{Manan H Anjaria*}
\author[1]{Mehmet Enes Erciyes*}
\author[1]{Vedant Ghatnekar}
\author[1]{Neha Navarkar}
\author[2]{Haritheja Etukuru}
\author[3]{Xiaole Jiang}
\author[1]{Kanad Patel}
\author[1]{Dhawal Kabra}
\author[1]{Nicholas Wojno}
\author[1]{Radhika Ajay Prayage}
\author[1]{Soumith Chintala}
\author[1]{Lerrel Pinto}
\author[2]{Nur Muhammad Mahi Shafiullah$^\dagger$}
\author[1]{Zichen Jeff Cui$^\dagger$}
\affil[1]{New York University}
\affil[2]{University of California, Berkeley}
\affil[3]{City University of New York}

\teaser{
\begin{figure}[h]
    \centering
    \includegraphics[width=0.8\textwidth]{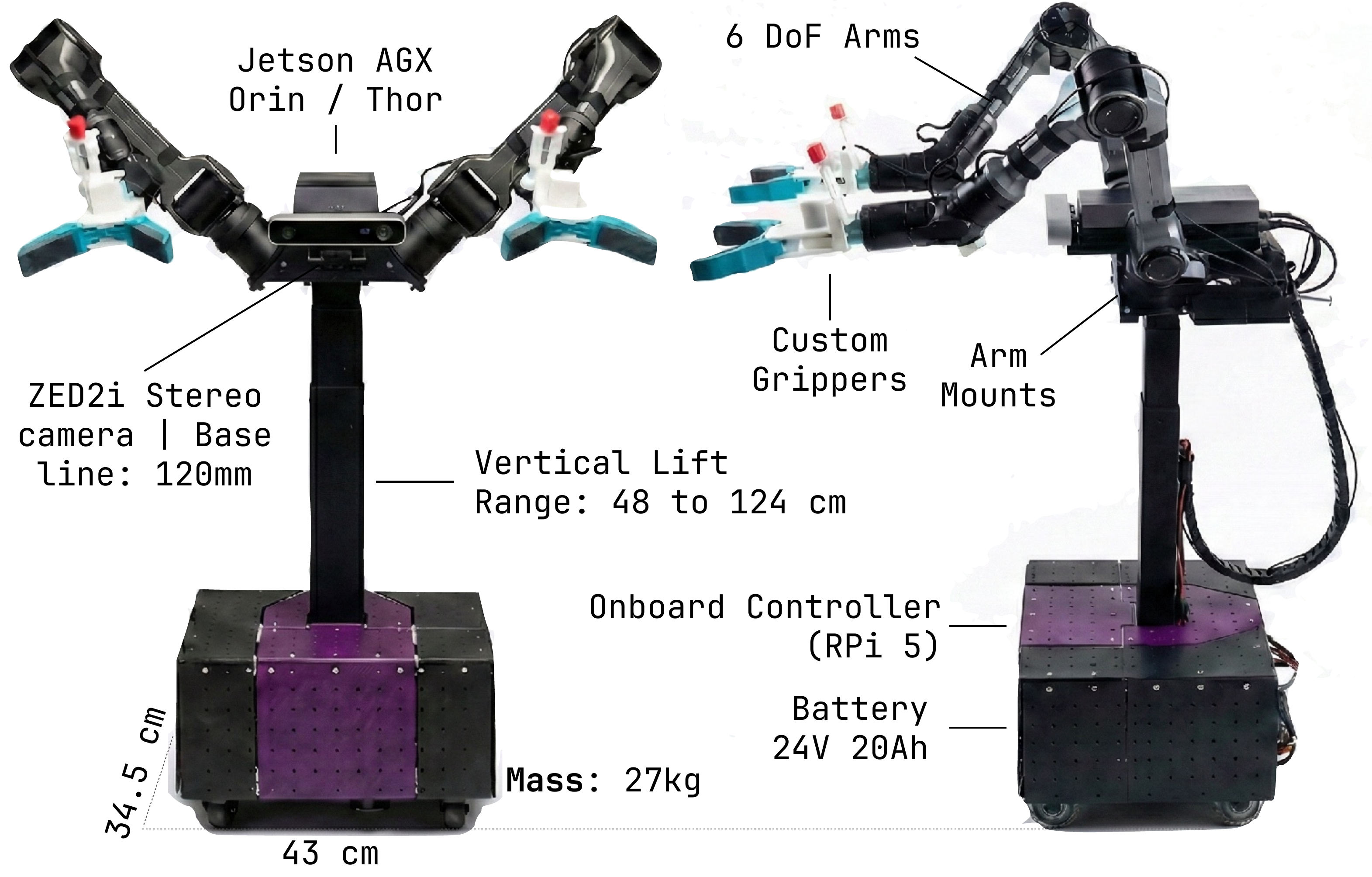}
    \caption{YOR is an open-source mobile manipulator combining an omnidirectional base, a lift, and two arms. YOR provides the dexterity and reach necessary for real-world tasks without the complexity of legged systems and the high cost of mobile manipulators on the market.}
    \label{fig:figure1}
\end{figure}
}

\begin{abstract}
Recent advances in robot learning have generated significant interest in capable platforms that may eventually approach human-level competence. This interest, combined with the commoditization of actuators, has propelled growth in low-cost robotic platforms. However, the optimal form factor for mobile manipulation, especially on a budget, remains an open question. We introduce YOR, an open-source, low-cost mobile manipulator that integrates an omnidirectional base, a telescopic vertical lift, and two arms with grippers to achieve whole-body mobility and manipulation. Our design emphasizes modularity, ease of assembly using off-the-shelf components, and affordability, with a bill-of-materials cost under US\$10,000. We demonstrate YOR's capability by completing tasks that require coordinated whole-body control, bimanual manipulation, and autonomous navigation. Overall, YOR offers competitive functionality for mobile manipulation research at a fraction of the cost of existing platforms.
\end{abstract}

\begin{document}

\setextrafooter{*Equal contribution\quad $^\dagger$Equal advising}
\maketitle

\section{Introduction}

The transformative power of robust hardware platforms and data collection infrastructure has fundamentally shaped recent advances in robotics. Dobb-E \citep{shafiullah2023bringing} and UMI \citep{chi2024universal} enabled portable data collection using hand-held grippers without requiring robot hardware during demonstration; ALOHA \citep{zhao2023aloha}, Open-TeleVision \citep{cheng2024opentelevisionteleoperationimmersiveactive} and OpenTeach \citep{iyer2024open} introduced intuitive bimanual teleoperation for stationary and mobile settings. Recently, models such as $\pi_{0.5}$ \citep{intelligence2025pi05visionlanguageactionmodelopenworld} and ACT-1 \citep{sunday_no_robot_data_2025} which scaled these ideas have demonstrated remarkable mobile manipulation capabilities. The success of the scaling efforts underscores a critical insight: the availability of capable yet affordable robot platforms is a key enabler for progress in the field. When platforms are expensive or difficult to maintain, research remains confined to laboratories, limiting the diversity of tasks, environments, and operational conditions that can be explored. Affordable and robust robotic platforms allow researchers to take them into the wild and run experiments that would otherwise be impossible.

Despite these advances, striking the right balance between manipulation capability, ease of use, and affordability has been a challenge. Humanoid robots, while promising, are still maturing: whole-body control and reliable locomotion remain active research challenges~\citep{he2024omnih2o, ze2025twist2}, making them less practical for large-scale mobile manipulation research today. Commercial mobile manipulators such as RB-Y1 \citep{rainbow2024rby1} and Dexmate \citep{dexmate2025} offer capable platforms, but come with significant drawbacks: high costs, large footprints unsuitable for home environments, and proprietary designs that are difficult to extend or customize for research purposes. TidyBot++~\citep{wu2023tidybot} demonstrated an affordable open-source alternative but relies on a single arm, lacking bimanual capabilities, and requires a long-reach manipulator to achieve sufficient workspace. XLeRobot \citep{wang2025xlerobot} offers an impressively low-cost entry point for robotics research, but its limited workspace, payload capacity, and durability constrain its applicability to real-world tasks. 

\begin{figure}[h!]
    \centering
    \includegraphics[width=\textwidth]{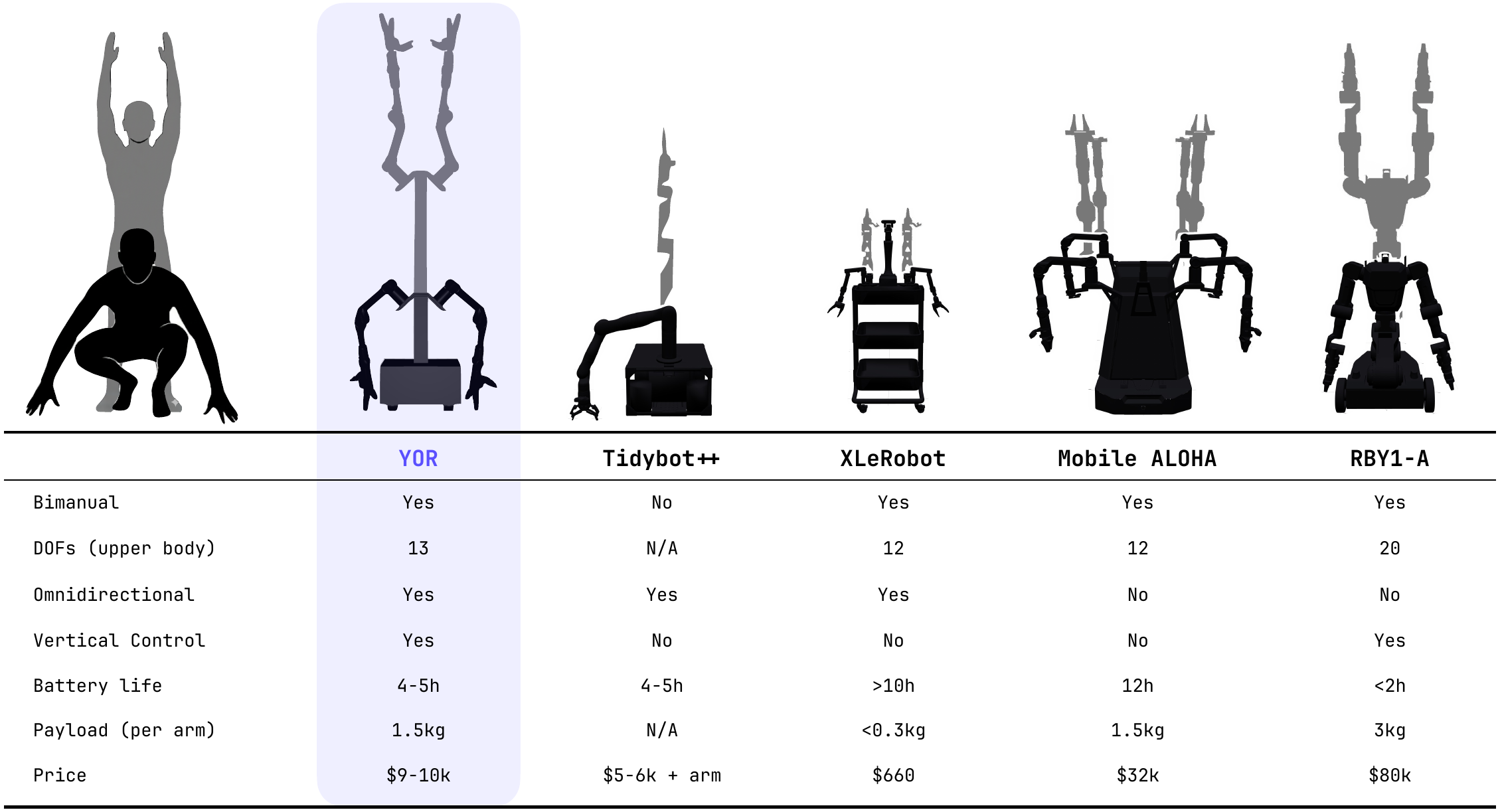}
    \caption{\textbf{Comparison of different mobile manipulators with YOR:} Tidybot++ \citep{wu2024tidybot++} relies on a single arm with no vertical control. XLeRobot \citep{wang2025xlerobot} is impressively low cost but its payload and reach limits its use cases. Due to its large footprint and non-holonomic base, Mobile ALOHA \citep{fu2024mobilealoha}  has limited navigation capability. RBY1-A \citep{rainbow2024rby1} and similar options on the market have many similar features to YOR but come with almost an order of magnitude higher price tag.}
    \label{fig:comparison}
\end{figure}

Based on these observations, we argue that an ideal platform for scalable mobile manipulation research should possess the following characteristics: (1) \textit{Low cost}, enabling practitioners on a budget to access capable robots; (2) \textit{Simple and robust control}, avoiding the complexity of balance controllers and minimizing hardware failures; (3) \textit{Omnidirectional mobility}, allowing intuitive navigation without non-holonomic constraints; (4) \textit{Bimanual manipulation with vertical reach}, providing the dexterity and workspace needed for real-world tasks; and (5) \textit{Compliance}, providing adaptability during manipulation and reducing the need for overly cautious and slow teleoperation.

We present YOR, an open-source mobile manipulator designed around these principles. YOR integrates an omnidirectional base, a telescopic vertical lift, and two compliant 6-DoF arms, achieving whole-body mobility and bimanual manipulation at a bill-of-materials cost under \$10K. YOR's swerve drive base features four independently rotating wheel modules, enabling omnidirectional motion that relaxes the non-holonomic constraints of traditional differential drive bases. With a footprint of $43$×$34.5$ cm, YOR can navigate cluttered indoor spaces where larger platforms cannot operate. A telescopic vertical lift with a stroke length of $63.5$ cm extends YOR's reach from floor level to overhead, providing the vertical workspace necessary for real-world household tasks such as picking objects from the ground or retrieving items from shelves. For bimanual manipulation, YOR is equipped with two PiPER arms fitted with custom grippers. The arms' compliant actuation ensures safe interaction during contact-rich tasks and reduces the need for overly cautious teleoperation. YOR integrates a ZED 2i stereo camera and supports onboard SLAM, enabling autonomous navigation and mapping in novel environments. YOR's slender profile allows an operator standing behind the robot to maintain direct line-of-sight to the workspace, facilitating intuitive teleoperation without relying on head-mounted displays or camera streaming setups.
Together, these design choices yield a platform that is low-cost, easy to build, and capable of whole-body bimanual manipulation and navigation. Our key contributions are as follows:

\begin{itemize}
    \item We identify a practical form factor for low-cost mobile manipulation research—an omnidirectional base with a vertical lift and dual compliant arms—that balances capability, controllability, and cost.
    \item We present YOR, a fully open-source hardware and software package that embodies this form factor, with a bill-of-materials cost of under US\$10K.
    \item We validate YOR's capabilities through integration tests demonstrating whole-body control, bimanual manipulation, and autonomous navigation.
\end{itemize}

\section{Related Work}

\subsection{Low-Cost Mobile Manipulation Platforms}
Most robotic hardware historically exists on a spectrum of precision and reliability, trading off against cost to find the optimal platform for any particular task. Early examples of this in mobile manipulators are Willow Garage's PR2~\citep{garage2012pr2} costing almost half a million US\$, and a decade later Hello Robot Stretch~\citep{kemp2022stretch} costing US\$20k. With the advent of learning-based control and cheap, compact actuators~\citep{katz2018low}, the precision requirements from hardware for coarse-grained mobile manipulation have also come down. Consequently, a series of low-cost hardware platforms have been built that rely primarily on learned policies, such as Mobile-ALOHA~\citep{fu2024mobile}, Tidybot++~\citep{wu2024tidybot++}, XLeRobot~\citep{wang2025xlerobot}, LeKiwi~\citep{lekiwi2025}, which we compare against in detail in Fig.~\ref{fig:comparison}. Their success has also inspired proprietary ventures in this space, in the form of RB-Y1~\citep{rainbow2024rby1}, DexMate~\citep{dexmate2025}, etc. Our work attempts to address the lack of an affordable yet capable robot with sufficient reach and mobility, and build upon prior work to find a novel form-factor at this price point.
We also note that with affordable mobile platforms, the prevalent interest is in mobile manipulation for indoor and home environments, as opposed to industrial applications ~\citep{shafiullah2023bringing,yenamandra2023homerobot,liu2024ok,liu2025dynamem,yang2024harmonic}. Therefore, our evaluation scenarios in this work also focus on indoor, home-like environments.

\subsection{Data Collection for Mobile Manipulation}
Recent works have proposed data collection platforms to address the limited availability of manipulation demonstrations. There are many systems which support fixed-arm setups such as OpenTeach~\citep{iyer2024open}, BunnyVisionPro~\citep{bunny-visionpro}, GELLO~\citep{wu2023gello}, ALOHA~\citep{zhao2023aloha}, and AirEXO~\citep{fang2023airexo}. Data collection for mobile manipulation presents additional challenges as the operator needs to control navigation and other degrees of freedom in the upper-body. Mobile ALOHA~\citep{fu2024mobilealoha} uses joint-space mappings between the leader and follower arms mounted on a mobile base that follows the operator's movements. This approach requires a base with a large footprint and cannot support control in vertical axis. JoyLo~\citep{jiang2025behavior} uses a GELLO setup with JoyCons attached at the ends, using buttons and joysticks to control the base and height. However, this setup requires the operator to be fixed in one place, making mobile teleoperation difficult. Tidybot++~\citep{wu2024tidybot++} proposes using pose tracking features and touch screens of mobile phones to collect data. While easy to set up and extend, using two mobile phones is not ergonomic and lack the tactile feedback of buttons and joysticks for prolonged use.

Several works proposed low-cost handheld data collection devices~\citep{etukuru2024rum, chi2024universal,choi2026inthewildcompliantmanipulationumift}. These devices enable scalable data collection without needing real robots and operators can record more dexterous demonstrations~\citep{generalist2025gen0}. Since robots need to learn from trajectories recorded by humans, mobile robots with more degrees of freedom are theoretically more capable of benefiting from this type of data and overcome kinematic limitations. Previous research~\citep{ha2024umilegsmakingmanipulation, gupta2025umionairembodimentawareguidanceembodimentagnostic} has explored how to deploy policies learned from this data on mobile platforms. However, integrating global navigation with local manipulation remains an challenge, particularly optimizing whole-body motion to handle environmental constraints and resolve kinematic redundancy.

\subsection{Mapping and Navigating Dynamic Environments}
Autonomous indoor navigation has been a foundational problem in robotics for over four decades, with early work by Moravec and Elfes \citep{Moravec-1985-15232} on occupancy grids. Classical algorithms remain central to modern systems: A* \citep{4082128} provides optimal graph-based path planning, while Pure Pursuit \citep{Coulter-1992-13338} enables robust trajectory tracking through geometric curvature commands. Modern commercial hardware \citep{keselman2017intelrealsensestereoscopicdepth,zed_sdk} and software packages such as OpenVINS \citep{geneva2020openvins}, cuVSLAM \citep{korovko2025cuvslam} allows easy integration of mapping capabilities to new platforms. Prior work such as \citep{bajracharya2024demonstrating} also inspires parts of our navigation stack.

\section{Hardware Design}

\begin{figure}[th!]
    \centering
    \includegraphics[width=0.7\linewidth]{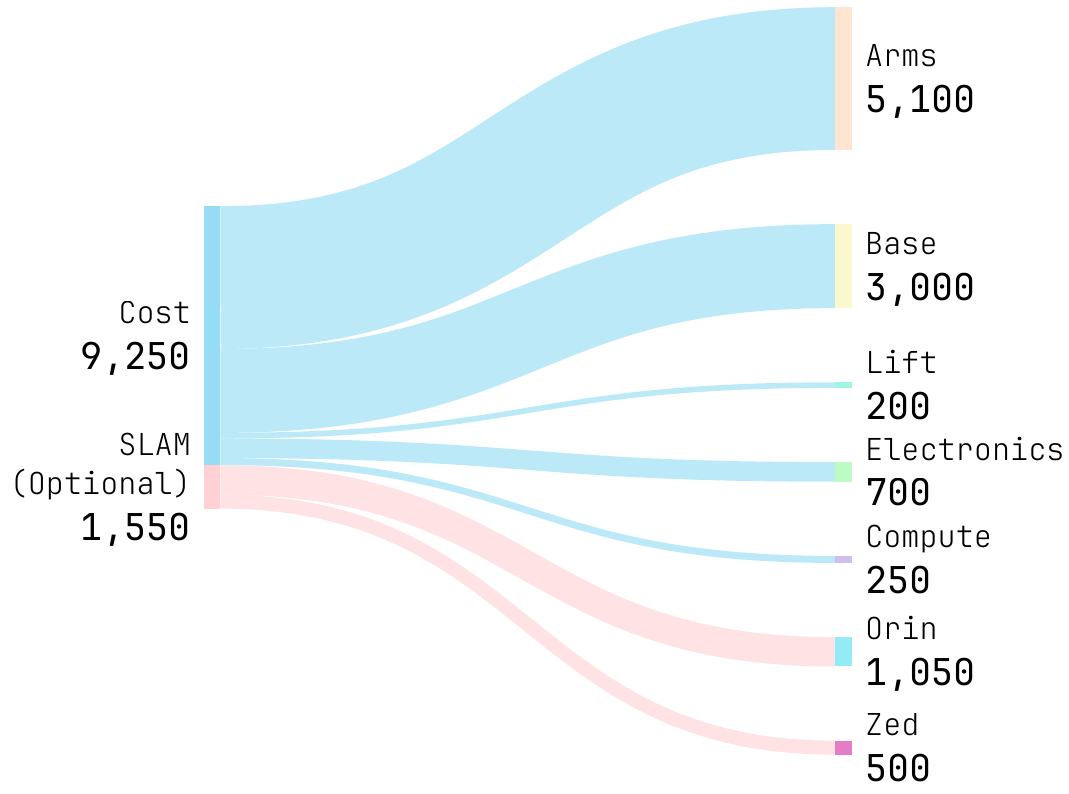}
    \caption{YOR is designed using off-the-shelf components in a modular configuration, with a total bill-of-materials cost of US\$9,250, the majority coming from the robot arms.}
    \label{fig:Cost Breakup}
\end{figure}


\subsection{Design Principles}

YOR is designed to have the most critical capabilities for mobile manipulation in human-centered environments with minimal complexity and cost. To achieve this, we adhere to four core principles:

\begin{itemize}
    \item \textbf{Low-Cost and Modularity:}
    To maximize accessibility, YOR is constructed from commercial off-the-shelf components. As illustrated in Fig. \ref{fig:Cost Breakup}, the total Bill of Materials (BOM) is approximately US\$9,250, significantly lower than comparable bimanual platforms. The modular design allows individual subsystems, such as the compute unit or the robotic arms, to be upgraded independently as hardware evolves.

    \item \textbf{Controllability:}
    Effective mobile manipulation requires precise positioning in cluttered environments. We prioritized a swerve drive base that allows YOR to reorient its manipulation workspace in any direction. YOR is also passively stable, which simplifies control during dynamic motion.

    \item \textbf{Safety and Compliance:}
    Designed for human-centered environments, YOR features a small footprint and low total mass to minimize kinetic energy. The robotic arms utilize compliant control strategies (detailed in Section \ref{subsec:Arms}) rather than stiff position control, ensuring safe interaction and adaptable manipulation. Additionally, the vertical lift mechanism is self-locking, preventing sudden drops in the event of power loss.

    \item \textbf{Open Source Accessibility:}
    To foster community development, the entire project is open-source. All CAD files, bill of materials, and software will be publicly shared, lowering the barrier to entry for researchers wanting to replicate or modify the platform for their specific needs.
\end{itemize}

\subsection{Mobile Base}
\label{subsec:base}

The mobility system of YOR is built upon an omnidirectional swerve drive architecture, enabling decoupled translational and rotational control. This agility is critical for mobile manipulation tasks, allowing the robot to precisely align its workspace with target objects in cluttered environments without the non-holonomic constraints of differential drive systems~\citep{Siegwart2011}.

\subsubsection{Hardware Configuration}
We utilize four commercial off-the-shelf REV Robotics $3$-inch MAXSwerve Modules \citep{REV_MaxSwerve} at a cost of approximately $\$625$ per module. Each module acts independently, integrating two brushless motors: a NEO $550$ coupled to an UltraPlanetary gearbox for azimuth (steering) control, and a high-torque NEO Vortex motor for wheel propulsion.

The modules feature a compact $3$-inch ($76.2$~mm) wheel diameter. This size is a deliberate design choice that minimizes the wheel's scrub radius, significantly reducing the static steering torque requirements compared to larger casters~\citep{Wong2008}. The compact footprint allows the wheels to be mounted at the absolute geometric extremities of the chassis, maximizing the area of the support polygon to improve stability.

\begin{figure}[th!]
    \centering
    \includegraphics[width=0.7\linewidth]{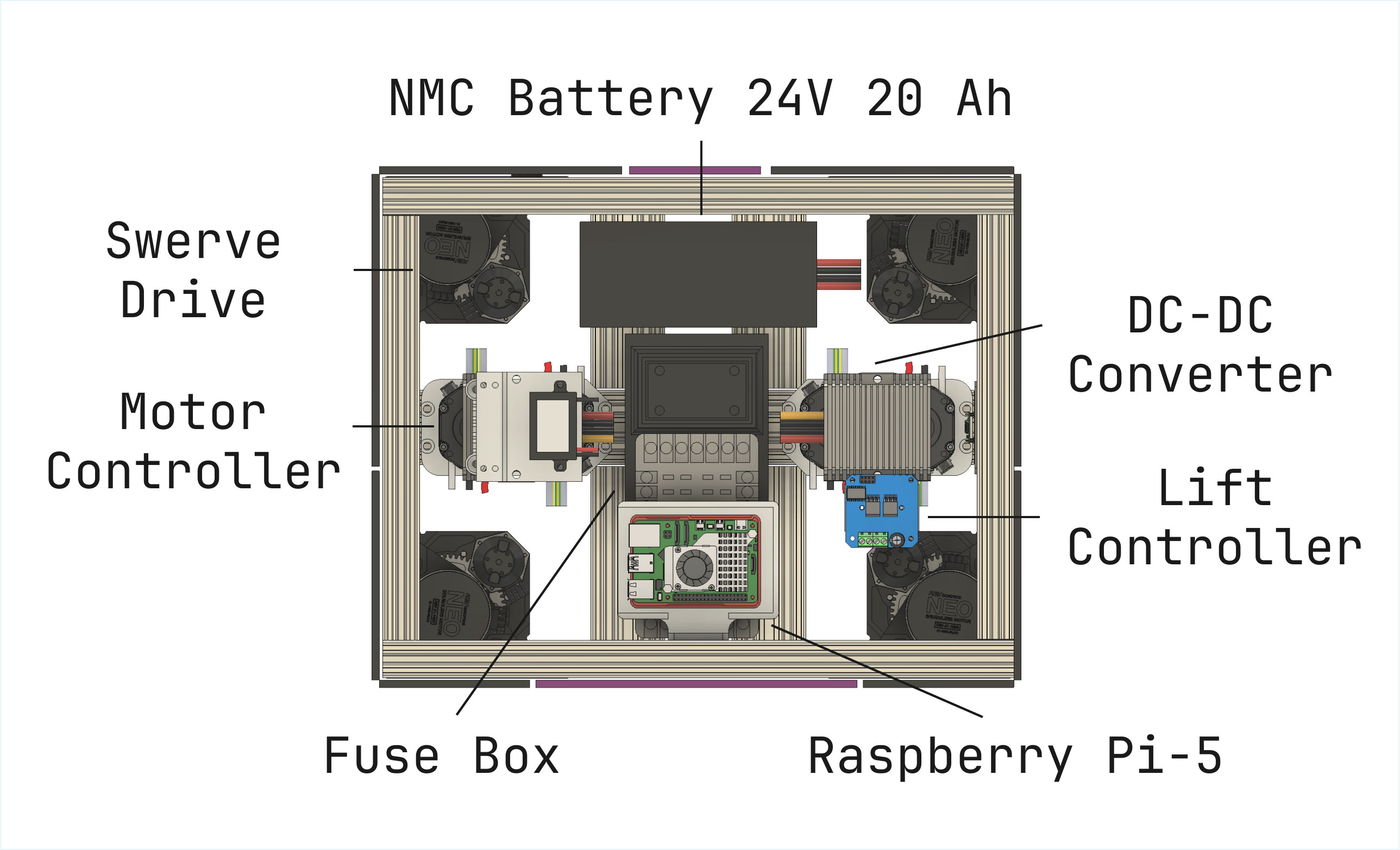}
    \caption{Internal layout of the mobile base following a ``heavy base'' philosophy. High-density components (battery, compute, and power distribution) are mounted centrally at the lowest elevation to lower the center of mass, flanked by four corner-mounted swerve modules.}
    \label{fig:Base}
\end{figure}

To support the dynamic loads of a vertical lift and dual manipulators, the base is constructed from dense aluminum extrusions and houses the high-density components—including the battery, power distribution, and primary compute units-at the lowest point as shown in Figure \ref{fig:Base}. This mass concentration lowers the global center of mass, maximizing the static stability margin and minimizing the risk of toppling during acceleration or heavy lifting~\citep{Messuri1985}.

\subsubsection{Kinematics and Control}
The robot's base frame $\mathcal{F}_b$ is defined at the geometric center. The kinematics mapping from the chassis velocity twist $\mathbf{v}_b = [v_x, v_y, \omega]^T$ to the individual wheel velocity vectors is governed by the coupling matrix $C \in \mathbb{R}^{8 \times 3}$~\citep{aoki2024switching}. Based on our chassis dimensions of half-width $W=0.152$~m and half-length $L=0.106$~m, the inverse kinematics relation is derived as:

\begin{equation}
\label{eq:swerve_ik}
\mathbf{V}_{\mathrm{modules}} = C \cdot \mathbf{v}_b =
\begin{bmatrix}
1 & 0 & W \\
1 & 0 & -W \\
1 & 0 & -W \\
1 & 0 & W \\
0 & 1 & L \\
0 & 1 & L \\
0 & 1 & -L \\
0 & 1 & -L
\end{bmatrix}
\begin{bmatrix}
v_x \\
v_y \\
\omega
\end{bmatrix}
\end{equation}

The resulting vector $\mathbf{V}_{\mathrm{modules}}$ contains the Cartesian velocity components for each of the four modules (FL, FR, RR, RL) organized as $\mathbf{V}_{\mathrm{modules}} = [v_{x1}, \dots, v_{x4}, v_{y1}, \dots, v_{y4}]^T$. These are converted to polar coordinates to generate the steering angle $\theta_i$ and drive velocity $v_i$ setpoints for the distributed motor controllers:
\begin{align}
\theta_i &= \operatorname{atan2}(v_{iy}, v_{ix}) \\
v_i &= \sqrt{v_{ix}^2 + v_{iy}^2}
\end{align}

To ensure smooth operation, we implement a ``shortest-turn'' optimization in the low-level controller. If a target angle $\theta_{target}$ requires a rotation greater than $90^\circ$ from the current state, the controller reverses the drive motor direction ($v_i \leftarrow -v_i$) and targets the equivalent opposite angle ($\theta_{target} \leftarrow \theta_{target} + \pi$). This minimizes steering latency and reduces wear on the azimuth gears~\citep{SwerveDriveKinematics}. 

Although the drive system is mechanically capable of achieving translational velocities up to $1.5$~m/s, we impose a software-defined velocity limit of $0.25$~m/s. This cap prioritizes safety during teleoperation and autonomous policy execution, ensuring the system maintains quasi-static stability even when the manipulators are fully extended or handling payloads.

\subsection{Vertical Lift}
\label{subsec:lift}

To extend the vertical workspace of the robot, YOR integrates a telescopic lift column originally designed for standing desks. We selected this commercial off-the-shelf mechanism for its proven reliability and mechanical rigidity; such actuators are typically validated for over $10,000$ load cycles and are engineered to maintain high stability under heavy offset loads, making them ideal for carrying the bimanual manipulator assembly~\citep{Wise2016}.

The lift provides a substantial stroke, capable of adjusting the shoulder height from a minimum of $0.60$~m ($23.6$ inches) to a maximum of $1.24$~m ($48.8$ inches). This wide range enables the robot to pick up objects off of the ground or high shelves, covering most of the relevant vertical workspace in human-centered environments.

For precise actuation, the lift employs a closed-loop control architecture managed by a dedicated Raspberry Pi Pico microcontroller \citep{raspberrypi_pico}. The Pico reads real-time position data from the lift's quadrature encoder and drives the DC motor via a BTS7960 high-current H-bridge~\citep{InfineonBTS7960}. This feedback loop allows for variable Pulse-Width Modulation (PWM) regulation, enabling accurate height positioning and stable velocity control up to a maximum of $35$~mm/s.

\subsection{Arms}
\label{subsec:Arms}
We build YOR as a bimanual robot that supports two 6-DoF arms and custom grippers as manipulation tools. We choose AgileX PiPER arms due to their low cost (US\$2,500) and light weight (4.2 kg). The arms are mounted onto a custom shoulder plate readily available from any online sheet metal service provider. This plate is designed to have arm mounting points at 45 degrees sideways tilt. We choose this shape to balance the forward and downward reach of the arms. The angled shoulders also prevent the elbows of the arms from colliding with each other even if their mounting points are close. Overall this design helps keep the robot footprint small.

For the end effector on the arms, we design custom grippers with an angular jaw that allows both precise manipulation and large force application. Each gripper uses an iPhone as an off-the-shelf sensor suite. 

As our primary focus is indoor household applications, we design YOR arms to be compliant. For operating around humans, a compliant controller is essential to adapt the manipulator's motion in response to external forces, and to ensure safety during deployment of a learned policy. Therefore, we implement a joint stiffness controller. The typical joint stiffness controller objective is \begin{align*}
\tau_{g}(q) + K_{p}(q_{\text{ref}} - q) + K_{d}(\dot{q}_{\text{ref}} - \dot{q})
\end{align*}

where $q$ is the measured joint positions and $q_{\text{ref}}$ is the target position set by the upstream controller. The system acts like a spring-damper around the reference position with stiffness coefficient $K_p$ and damping coefficient $K_d$. The feedforward torque gravity compensation allows us to set low stiffness gains, resulting in compliant movement. This arm controller runs with two layers. The low-level real time controller runs at 200 Hz. It uses Ruckig \citep{berscheid2021jerk} real time trajectory generation to make sure that the commands obey safety limits. The high-level controller sets targets for the low-level controller to track at lower frequencies. We provide both joint control and task-space control for the arms. In joint control mode, the targets are set directly. In task-space control mode, Mink~\citep{mink} is used to solve inverse kinematics.

\section{Applications of YOR}
\subsection{Whole-Body Teleoperation}
\begin{figure}[h!]
    \centering
    \includegraphics[width=\linewidth]{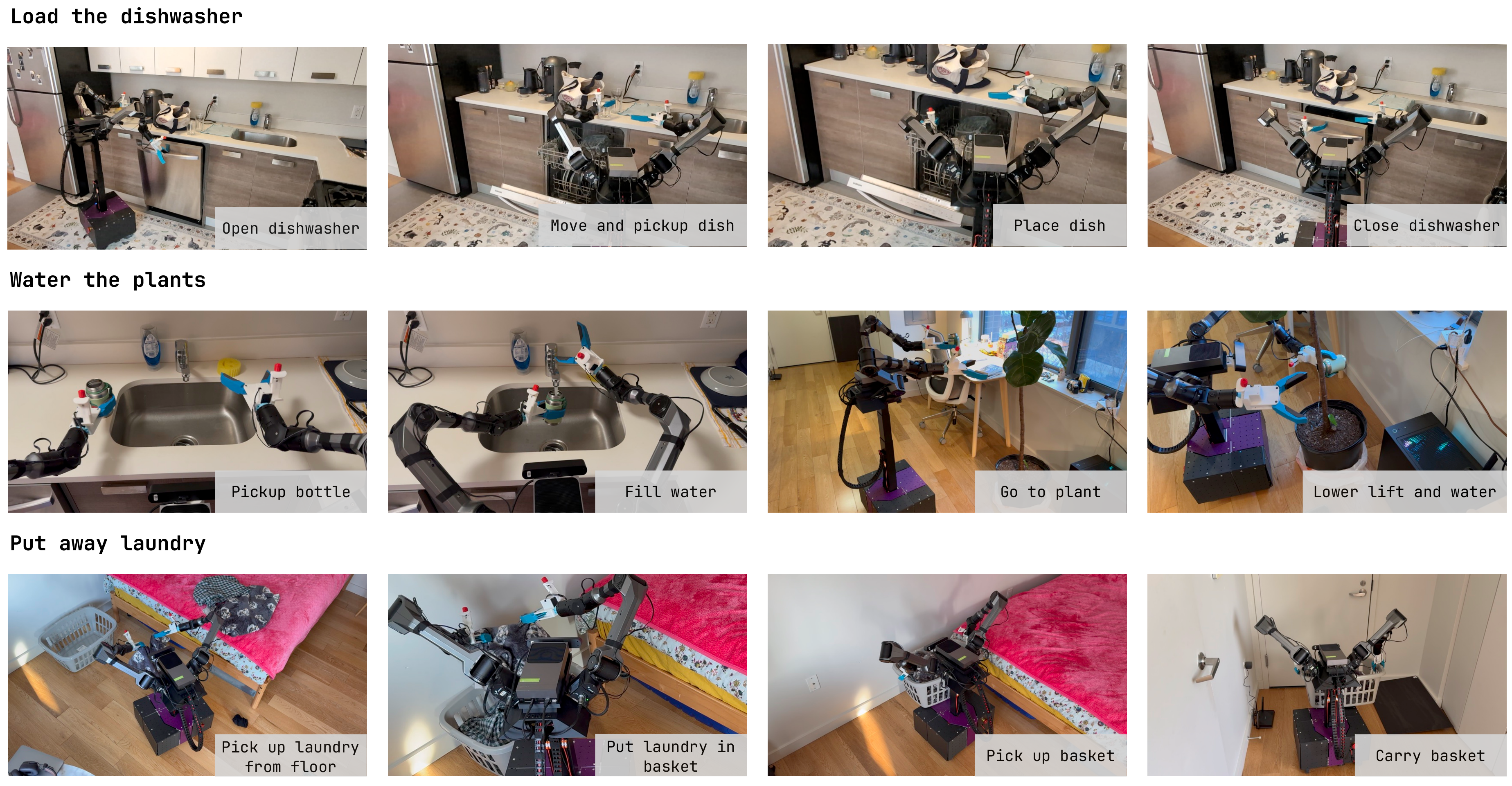}
    \caption{A teleoperated YOR completes daily tasks in home environments with its extended workspace and bimanual arms. In these tasks, we teleoperate YOR to open a dishwasher and use its mobile base to load and close it. YOR can also use the lift to lower its height and water plants or pick up a basket from the ground.}
    \label{fig:teleop_examples}
\end{figure}

To enable imitation learning on our platform, we design an intuitive whole-body teleoperation system that uses Meta Quest 3/3S controllers to control all components of YOR. The pose of the Quest controllers are calibrated and retargeted to the end-effector (EE) pose of the arms. Instead of retargeting the operator's bending motions to control height, we choose to directly control the lift via sending \texttt{[up, down]} velocity command with the grip buttons on the controllers. The inverse kinematics task is defined as the relative pose of end-effector with respect to top of the lift. Compared to whole-body retargeting, this design choice eliminates the need for the operator to squat in uncomfortable positions and decouple the height and arm control. To control the base, we use the joysticks on the controllers to command linear and angular velocities. In Figure \ref{fig:teleop_examples}, we show some examples of tasks YOR is able to complete with this teleoperation system. 

\subsection{Policy Learning}

\begin{figure}[th!]
    \centering
    \includegraphics[width=\linewidth]{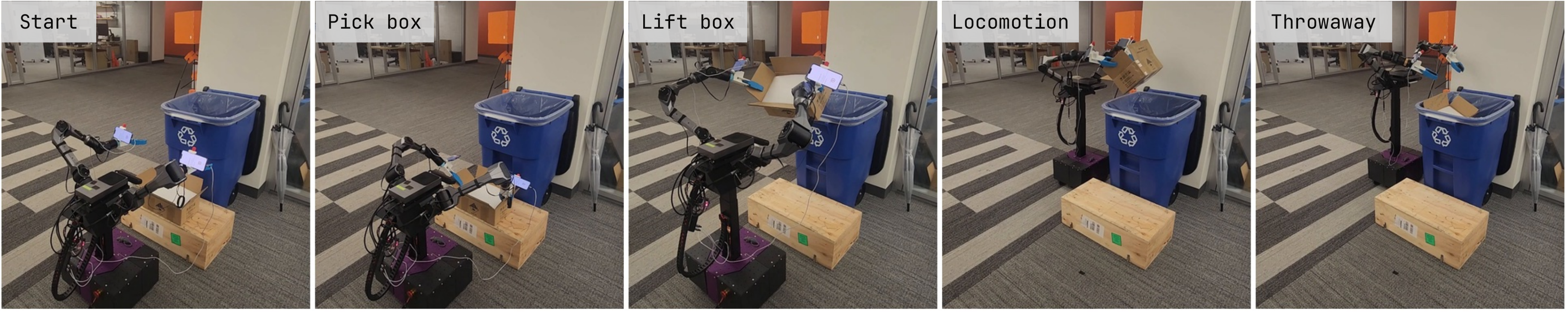}
    \caption{We evaluate YOR's bimanual whole-body locomanipulation capability through an imitation learning policy. In this task, YOR can autonomously pick up a box, navigate around an obstacle and drop it in the recycling bin.}
    \label{fig:policy_rollout}
\end{figure}
In this section, we demonstrate YOR's capacity as a platform for bimanual mobile manipulation policy learning. We train YOR on a ``recycling'' task, throwing away a large cardboard box into a recycle bin. This task requires YOR to pick up the box with both grippers, lift the box up and over the recycle bin, move around an obstacle to approach the bin, and place the box inside. See Figure \ref{fig:policy_rollout} for visualization.

\subsubsection{Data collection} We collect expert demonstrations with teleoperation at 30Hz and discard any trajectories where the robot loses odometry tracking, yielding a final dataset of 100 trajectories. For visual input, we record $256 \times 256$ RGB images from two iPhone wrist cameras following the setup in \citep{etukuru2024rum} and the ZED 2i head camera. For proprioception, we record end-effector poses in the top-of-lift reference frame, lift height, and base odometry from ZED. We set the first frame odometry as the world frame of that whole trajectory. The policy predicts a 19-dimensional action (7D end-effector pose and gripper command per arm, base translation, $\sin, \cos$ of base yaw and lift position). During preprocessing, we drop static frames from each trajectory.

\subsubsection{Architecture}
We adopt VQ-BeT \citep{lee2024behavior} for policy learning. The wrist and head camera image features are extracted via a shared ResNet-50 encoder trained using self-supervised learning on a diverse set of pickup data; each image feature is then concatenated with the linearly up-projected proprioception data. In total, at each timestep, we have 4 new observation tokens each of which consists of 768 dimension of visual feature and 256 dimension of proprioception, where each wrist contributes an observation token and the head camera image is split into two observation tokens from left/right crop. A transformer decoder then predicts discretized actions and action offsets from a window of past three observations. We refer readers to \citet{lee2024behavior} for more architectural details.

\subsubsection{Evaluation}
The onboard Jetson computes and publishes robot pose in real time, and ZED image and wrist camera images are all published via ZMQ \citep{cui_jeffaccecommlink_2026} in real time for remote machine inference. We stream the policy action to the Raspberry Pi onboard, which controls both the arms, the lift, and communicates with Jetson through a local network to run base position PID control. We deploy the trained policy for 10 trials as a proof of concept to validate our hardware capabilities.

\begin{center}
\begin{tabular}{ ccccc } 
 \hline
 & Pick Up & Lifting & Locomotion & Total \\ 
 \hline
 Success Rate & 10/10 & 10/10 & 9/10 & 9/10 \\  
 \hline
\end{tabular}
\end{center}
The policy is able to learn to grab the box with both grippers, and lift it up to the desired height. The main failure mode was odometry drift due to head camera occlusion.

\subsection{SLAM and Navigation} \label{sec:SLAM}
\begin{figure}[htb!]
    \centering
    \includegraphics[width=\linewidth]{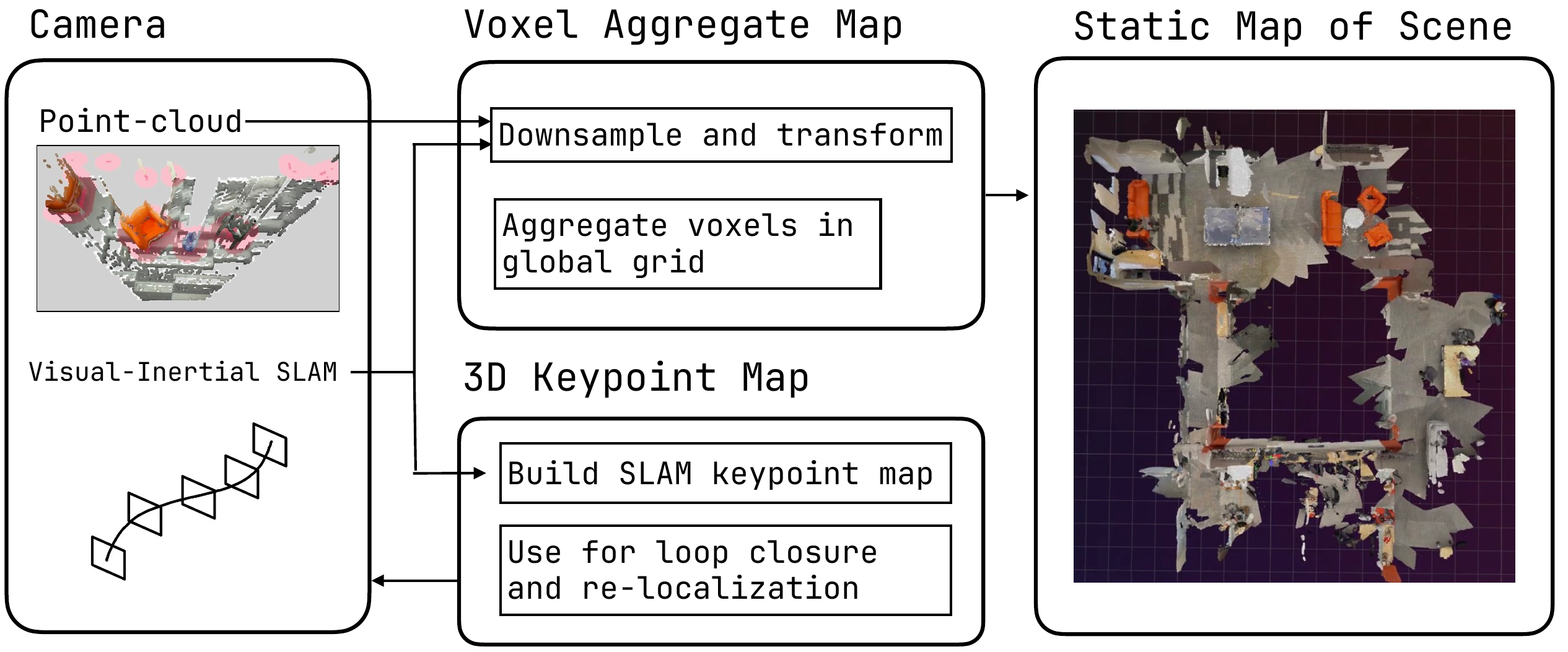}
    \caption{YOR's mapping pipeline combines the point clouds and visual-intertial odometry provided by the head-mounted ZED camera.}
    \label{fig:slam_mapping}
\end{figure}
\subsubsection{Motivation}
Whole-body mobile manipulation requires accurate base localization within a geometric model of the environment to reach manipulation-relevant poses while avoiding collisions in clutter. Our system therefore couples real-time state estimation with dense mapping and classical planning to provide a reliable, metric backbone for task execution.

\subsubsection{Architecture}
Onboard the robot, we use visual--inertial SLAM~\citep{qin2018vins} to estimate the robot pose as a time-varying rigid transform $T_{\text{WB}}(t)\in \text{SE}(3)$. This transform is used for closed-loop control on pose and for map-fusion in a consistent world-frame. Our perception stack is based on the ZED 2i stereo depth camera. Odometry comes from the ZED-SDK~\citep{zed_sdk}, along with loop closure signals, which are used to compute the final pose of the camera in the world frame. RGB and stereo depth observations are converted to point clouds and integrated into a voxel map in the world frame, with voxel filtering and outlier rejection at 5 Hz. For navigation, we employ histogram based floor detection to find the lowest height mode in the voxel data, then project the rest of the voxels onto a 2D grid. We derive the global map after sufficient voxel inflation to account for robot size and additional voxel inflation to add higher cost near obstacles. The live point cloud is down-sampled, transformed to the world frame and similarly projected to a local 2D cost map. This is integrated with the global cost map at 10 Hz by linearly interpolating the cells. Mapping updates are gated by pose-quality checks to prevent map corruption during tracking degradation. We plan collision-free paths using a weighted A* planner and plan again if new obstacles block the current path. The resulting waypoint sequence is sent to the base controller which uses Pure Pursuit~\citep{coulter1992implementation} algorithm to track them. A PID controller is used to track a look-ahead point along this path at 50 Hz with feedback from the estimated base pose. We also similarly provide a non-planner based absolute and delta position controller for policy based control. State estimation, mapping, planning, and control modules are run in separate processes and communicate via a lightweight RPC solution \citep{cui_jeffaccecommlink_2026}. This design provides real-time performance and modularity for customization.

\begin{figure}[t!]
    \centering
    \includegraphics[width=1\linewidth]{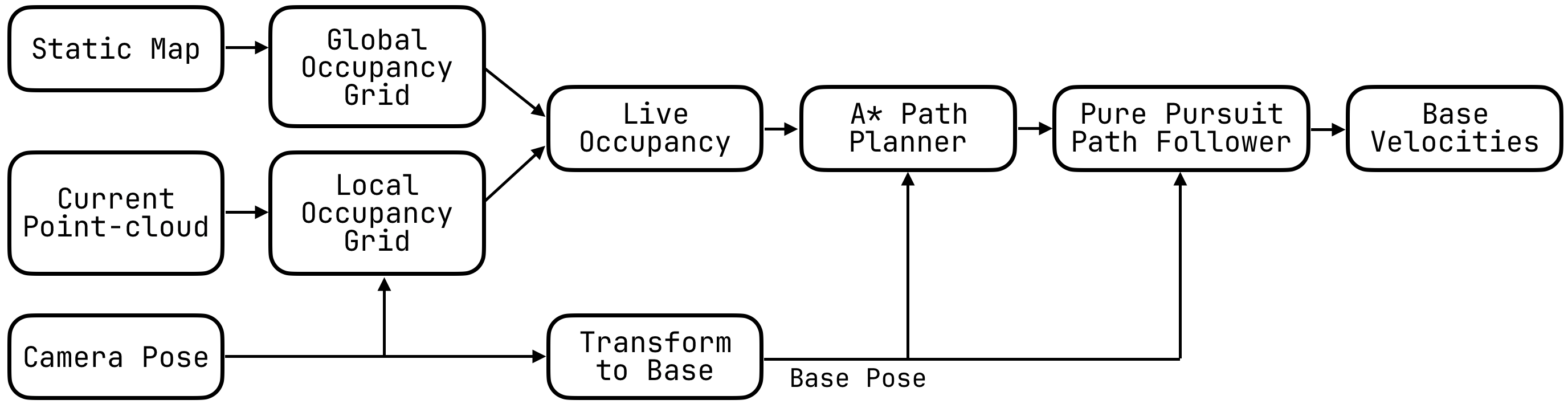}
    \caption{Using the static map and current observations, YOR can dynamically plan a path, avoid obstacles and accomplish indoor navigation.}
    \label{fig:slam_planning_nav}
\end{figure}

\subsubsection{Experiments}

\begin{figure}[th!]
\centering
    \includegraphics[width=0.5\linewidth]{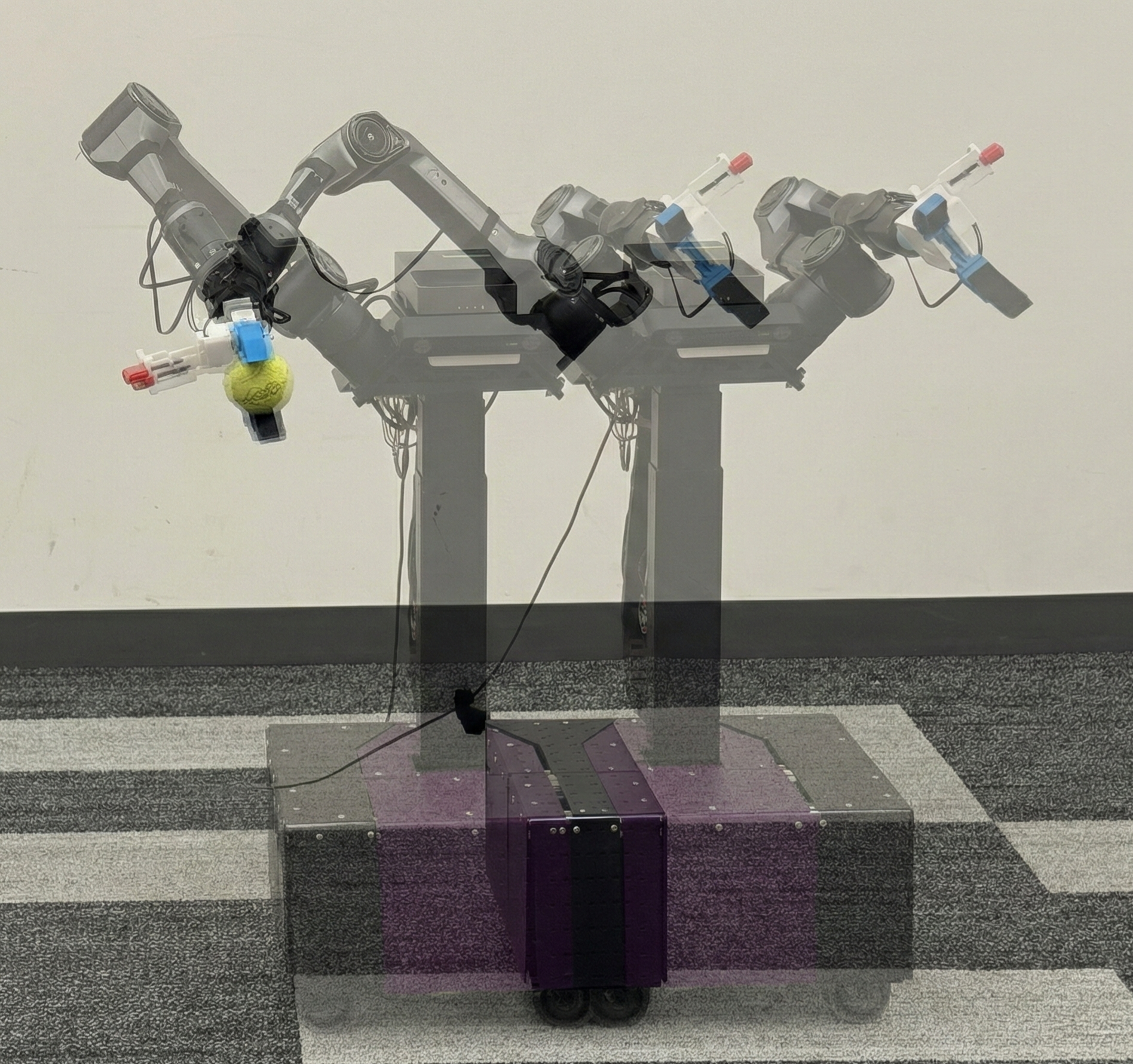}
    \caption{Demonstration of whole-body coordination. The arm controller actively compensates for base motion in real-time, effectively locking the end-effector's pose in the world frame while YOR translates and rotates around it.}
    \label{fig:chicken_head}
\end{figure}
\begin{figure}[th!]
\centering
    \includegraphics[width=0.67\linewidth]{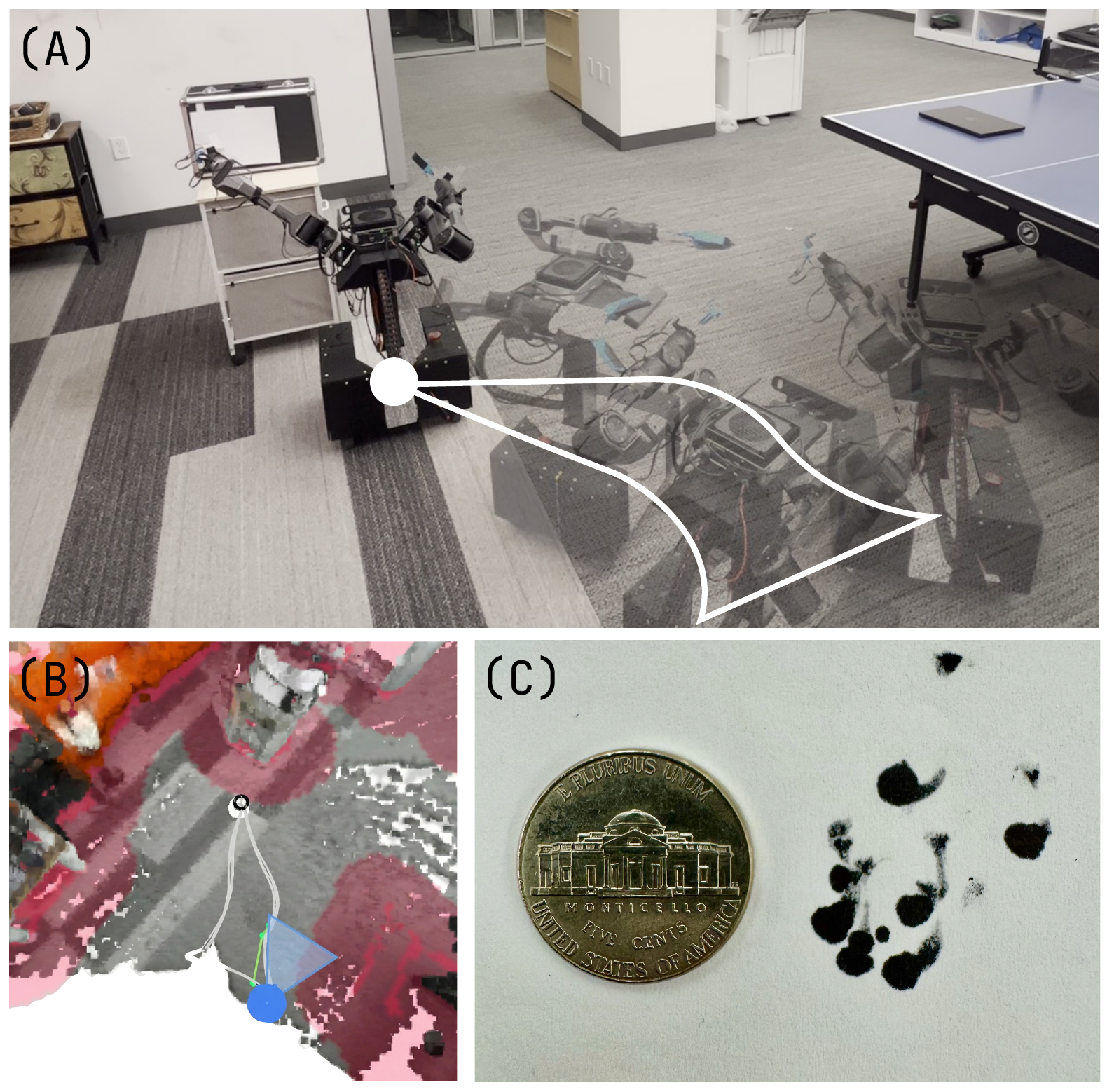}
    \caption{To display YOR's precision in mobile manipulation, the robot moves in a loop and make marks at the same position on a small paper in (A). In (B) we show the target loop in the robot's internal representation, and in (C) we display the resultant marks on the paper. Next to the marks is a US nickel (radius 10.6 mm) for scale.}
    \label{fig:Tally Mark}
\end{figure}

\begin{figure}[th!]
    \centering
    \includegraphics[width=\linewidth]{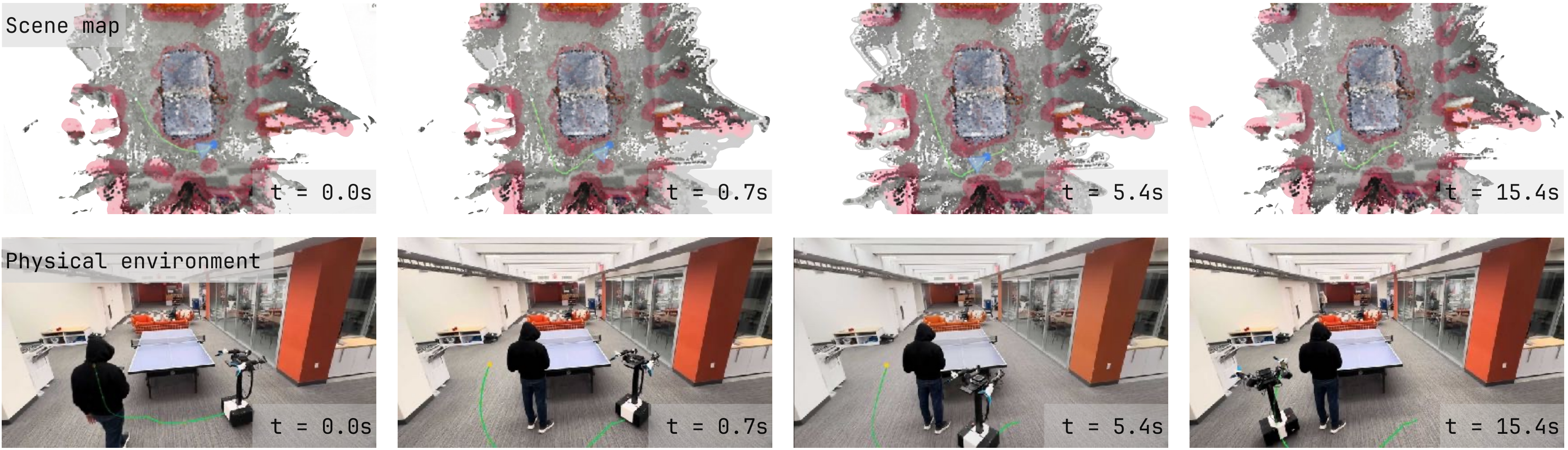}
    \caption{Real-time obstacle avoidance in YOR's SLAM systems. When YOR detects a novel obstacle (in this case, a walking human), it plans a new path around them within a second to reach the goal point without collision. On the top, we show the internal voxel map representation, where Green shows the planned path and red shows detected obstacles. The bottom row shows third-person view of the environment and the plan.}
    \label{fig:obstacle_avoidance}
\end{figure}

We demonstrate the whole-body control, repeatability, accuracy, stability, real-time planning and navigation of the SLAM system using tally-mark, whole-body coordination and dynamic obstacle avoidance demos.

\paragraph{Odometry and loop closure} We demonstrate YOR's odometry accuracy and loop-closure performance via the tally-mark demo. As illustrated in Fig.~\ref{fig:Tally Mark}, YOR marked a reference point on a white paper and returned to mark the sheet once every loop, while traversing the environment. After ten loops, the return marks were within a radius of  12 mm from the initial mark, which is within the positional accuracy (50 mm) of our SLAM pipeline.

\paragraph{Whole-body coordination} The whole-body coordination demo shows the accuracy of the base pose estimation and the end effector task space control coupled together during diverse motions. To illustrate this, YOR held a ball in its gripper while the mobile base executed a lateral translation, with the arm controller compensating for the base movement. As shown in the Fig. \ref{fig:chicken_head}, the end-effector deviated by 16 mm, showcasing minimal tracking error of the whole-body control system.

\paragraph{Dynamic obstacle avoidance} In the  dynamic obstacle avoidance demo, YOR navigates toward a fixed goal while a walking human interrupts the planned route. Upon detection, the planning algorithm updates the voxel occupancy and triggers a replan, producing a collision-free path around the human within one second. Fig.~\ref{fig:obstacle_avoidance} visualizes this behavior: the top row shows the internal voxel-map representation with the planned path in green and detected obstacles in red, while the bottom row shows the corresponding third-person view and the resulting plan in the physical environment.

\section{Conclusion}
\label{sec:conclusion}
In this work, we introduce YOR, an open source, affordable bimanual mobile manipulator. While in its current form YOR provides many attractive capacities for researchers in mobile manipulation, there are further research in the future that will make this platform even stronger. One straightforward improvement would be to mount affordable 7-DOF arms~\citep{enactic2025openarm} instead of 6-DOF ones to decrease the possibility of singularities while teleoperating or deploying policies on YOR. Similarly, while we have integrated capable SLAM systems onto YOR, state-of-the-art in semantic navigation~\citep{liu2025dynamem} would make it even more capable in performing zero-shot mobile manipulation in the real world.
Finally, further research into whole-body controllers and perceptive loco-manipulation will also open up the possibility of many new dynamic and dexterous tasks that may be difficult with our current setup.

\bibliographystyle{yearnat}
\bibliography{references}

\clearpage
\appendix
\section{Appendix}
\subsection{Bill of materials}

YOR is made of modular components that are available off-the-shelf on the market. Table \ref{table:bom} shows a breakdown of each component necessary to build YOR and their current price. 

\vspace{-1em}
\subsection{Policy learning hyperparameters}
\begin{center}
\begin{tabular}{cc}
    \toprule
     Observation context & 3 \\
     Action chunk length & 10 \\ 
     VQ-BeT codebook size & 8 \\
     VQ-BeT embed dim & 512 \\ 
     VQ-BeT depth & 6 \\ 
     VQ-BeT head & 4 \\ 
     Optimizer & AdamW \\
     Learning rate & 6e-4 \\
     Batch size & 512 \\
    \bottomrule
\end{tabular}
\end{center}

\vspace{-1em}
\subsection{SLAM and navigation details}

\begin{table}[h!]
\centering
\footnotesize
\begin{tabular}{lc}
    \toprule
    \textbf{Parameter} & \textbf{Value} \\
    \midrule
    \multicolumn{2}{l}{\textbf{Mapping Params}} \\
    Global map voxel size & 0.02 m\\
    Outlier rejection radius & 0.12 m\\
    Outlier rejection neighbors & 3 \\
    Floor band & 0.1 m \\
    
    \midrule
    \multicolumn{2}{l}{\textbf{Occupancy Grid Params}} \\
    Cell size & 0.05 m \\
    Floor band & 0.25 m\\
    Robot inflation radius & 0.3 m \\
    \midrule
    \multicolumn{2}{l}{\textbf{Base Controller Params}} \\
    \multicolumn{2}{l}{\emph{Pure Pursuit}} \\
    Look-ahead distance range & 0.2 - 0.4 m \\
    Velocity smoothing factor & 0.2 \\
    \multicolumn{2}{l}{\emph{Position PID}} \\
    Position: kp & 1.5 \\
    Position: ki & 0.02 \\
    Position: kd & 0.15 \\
    Yaw: kp & 2.1 \\
    Yaw: ki & 0.01 \\
    Yaw: kd & 0.2 \\
    Position Tolerance & 0.015 m\\
    Yaw Tolerance & 0.03 rad\\
    Velocity max & 0.35 m/s\\
    Omega max & 1.0 rad/s\\
    \bottomrule
\end{tabular}
\end{table}

\subsubsection{Base Position Control PID Tuning}
We tune the base position controller gains to ensure convergence within the specified position and yaw tolerances. Tuning proceeds by first applying constant translational and rotational setpoints and adjusting the proportional gains to achieve a desirable convergence rate. We then increase the derivative gains to suppress overshoot and finally adjust the integral gains to compensate for near-setpoint ground friction and residual steady-state error.

\subsubsection{Velocity Filtering}
The base controller smooths velocity commands with an exponential
moving average (EMA) filter:
\begin{equation}
  \mathbf{v}_{\mathrm{filt}}(t)
    = (1-\alpha)\,\mathbf{v}_{\mathrm{cmd}}(t)
    + \alpha\,\mathbf{v}_{\mathrm{filt}}(t-1),
\end{equation}
where $\mathbf{v}_{\mathrm{cmd}}(t)$ is the commanded velocity,
$\mathbf{v}_{\mathrm{filt}}(t-1)$ the previous filtered value, and
$\alpha=0.2$.

\subsubsection{Floor Filtering}
The floor height is estimated by computing the 5th percentile of the
vertical point-cloud coordinates, averaging all points within a
$0.1\,\mathrm{m}$ band above this percentile, and temporally smoothing
the result with an EMA.

\subsubsection{Odometry and Loop-Closure Experiment}
The tally-mark demonstration validates YOR's loop-closure capability.
A ZED-camera SLAM pipeline estimates the base pose and builds a static
occupancy grid for A* path planning. Planned paths are streamed to the
robot via RPC with waypoint spacing of roughly $2\times$ the grid cell
size ($0.05$--$0.15\,\mathrm{m}$); planning terminates once the robot
is within $0.02\,\mathrm{m}$ of the goal.

Each closed-loop \emph{tally} follows the route
HOME$\,\rightarrow\,$P1$\,\rightarrow\,$P2$\,\rightarrow\,$P3$\,\rightarrow\,$HOME.
Upon return, a two-stage docking procedure is executed: an absolute
$\mathrm{move\_to}(x,z,\psi)$ with $2\,\mathrm{cm}$ translational
tolerance, followed by in-place heading alignment to the saved HOME
direction with $0.04\,\mathrm{rad}$ tolerance. End-of-loop drift is
reported as $(\Delta x,\Delta z)$ and yaw error, where
$\psi=0$ faces $+Z$ and $\psi=\pi/2$ faces $+X$. At HOME the arm is
commanded to a fixed \emph{mark} configuration.

\subsubsection{Whole-body coordination experiment}
We evaluate whole-body coordination by keeping the end-effector (EE) fixed in
the world frame while the mobile base moves. The system consists of two
concurrent processes: base localization and whole-body control. The
localization node streams the base pose
$[q_x, q_y, q_z, q_w,\; x, y, z]$ with a timestamp at 120\,fps. On the
control side, a non-blocking receiver thread caches the most recent pose, and
the main control loop runs at 120\,Hz.

At the start of each trial we record the initial world-to-base transform from
the localizer and the initial base-to-EE transform from the robot interface;
together these define the EE's initial world-frame pose. At every control
step the commanded EE pose is updated so that this world-frame pose remains
invariant under base motion---that is, translation and yaw changes reported
by the base localization are compensated by the arm. To reduce visible jitter while
preserving responsiveness, we apply lightweight smoothing to the commanded EE
pose: an exponential low-pass filter on translation with time constant
$\tau_{\mathrm{trans}} = 0.20\,\mathrm{s}$ and SLERP-based smoothing on
orientation with $\tau_{\mathrm{rot}} = 0.30\,\mathrm{s}$, both using the
blending coefficient $\alpha = 1 - \exp(-\Delta t / \tau)$.

\begin{table}[ht!]
\vspace{-1em}
\centering
\footnotesize
\begin{tabular}{lccc}
\textbf{Item} & \textbf{Count} & \textbf{Price (USD)} & \textbf{Cumulative Price (USD)} \\
\midrule
\multicolumn{4}{l}{\textbf{Arms}} \\
Piper Arms & 2 & 2,500.00 & 5,000.00 \\
Dynamixels Motors & 2 & 27.00 & 54.00 \\
U2D2 & 1 & 36.92 & 36.92 \\
U2D2 Power Hub Board & 1 & 21.85 & 21.85 \\
Arms Attachment Plate & 1 & 76.74 & 76.74 \\
USB Hub & 1 & 9.99 & 9.99 \\
\midrule
\multicolumn{4}{l}{\textbf{Base (NEO Vortex)}} \\
3 inch Max Swerve Module & 4 & 675.00 & 2,700.00 \\
\midrule
\multicolumn{4}{l}{\textbf{Aluminum Extrusions}} \\
Double Six Slot Rail, 6.5" & 2 & 7.28 & 14.56 \\
Double Six Slot Rail, 10" & 2 & 11.20 & 22.40 \\
Double Six Slot Rail, 15" & 2 & 16.80 & 33.60 \\
Double Six Slot Rail, 4.75" & 8 & 5.32 & 42.56 \\
Single Four Slot Rail, 17" & 2 & 26.35 & 52.70 \\
Single Four Slot Rail, 11.5" & 2 & 17.83 & 35.66 \\
90 Degree Clamp & 1 & 20.88 & 20.88 \\
L Shape Interior Connector & 1 & 26.90 & 26.90 \\
T-Clamp & 1 & 19.99 & 19.99 \\
\midrule
\multicolumn{4}{l}{\textbf{Lift}} \\
Lift & 1 & 165.00 & 165.00 \\
Lift Controller & 1 & 10.00 & 10.00 \\
\midrule
\multicolumn{4}{l}{\textbf{Power System}} \\
Battery + Charger & 1 & 350.00 & 350.00 \\
24V--24V 6A Regulator & 2 & 19.00 & 38.00 \\
24V--5V 5A Regulator & 1 & 14.00 & 14.00 \\
24V--12V Regulator & 1 & 19.00 & 19.00 \\
Power Distribution Board & 1 & 16.00 & 16.00 \\
CANbus Module & 1 & 20.00 & 20.00 \\
Emergency Stop Switch & 1 & 17.00 & 17.00 \\
\midrule
\multicolumn{4}{l}{\textbf{Compute Module}} \\
Raspberry Pi 5 (16 GB) & 1 & 144.50 & 144.50 \\
Raspberry Pi 5 Active Cooler & 1 & 9.99 & 9.99 \\
USB Hub & 1 & 9.99 & 9.99 \\
Wireless Gamepad & 1 & 39.00 & 39.00 \\
Raspberry Pi Pico & 1 & 7.34 & 7.34 \\
Pico Breakout Board & 1 & 11.95 & 11.95 \\
Cable Carrier & 1 & 14.00 & 14.00 \\
USB Type-C Cable & 1 & 9.00 & 9.00 \\
CAT Cable & 1 & 7.00 & 7.00 \\
\midrule
\multicolumn{4}{l}{\textbf{Miscellaneous}} \\
Wago Splice Box (60 pcs) & 1 & 27.00 & 27.00 \\
Wago 2-Port Connector (50 pcs) & 1 & 35.00 & 35.00 \\
Wire Crimp 10--12 AWG & 1 & 10.00 & 10.00 \\
Wire Crimp 16--22 AWG & 1 & 10.00 & 10.00 \\
Terminal Block & 1 & 8.00 & 8.00 \\
Barrel Connector & 1 & 4.00 & 4.00 \\
\midrule
\multicolumn{4}{l}{\textbf{Nuts and Bolts}} \\
M5 x 10 mm & 1 & 9.00 & 9.00 \\
M5 Extrusion Nut & 1 & 15.00 & 15.00 \\
10--32 Screws & 1 & 9.00 & 9.00 \\
Anderson Battery Connector & 1 & 10.00 & 10.00 \\
\midrule
\multicolumn{3}{r}{\textbf{Total}} & \textbf{9,207.52} \\
\end{tabular}
\caption{Bill of Materials and Cost Breakdown.}
\label{table:bom}
\end{table}

\end{document}